%% file: main.tex
\documentclass[10pt,twocolumn,letterpaper]{article}

\usepackage{wacv}              

\usepackage{graphicx}
\usepackage{amsmath}
\usepackage{amssymb}
\usepackage{booktabs}
\usepackage[accsupp]{axessibility}

\input{preamble}

\usepackage[pagebackref,breaklinks,colorlinks]{hyperref}

\usepackage[capitalize]{cleveref}
\crefname{section}{Sec.}{Secs.}
\Crefname{section}{Section}{Sections}
\Crefname{table}{Table}{Tables}
\crefname{table}{Tab.}{Tabs.}

\begin{document}


\title{EdgeGaussians - 3D Edge Mapping via Gaussian Splatting}

\author{
Kunal Chelani$^1$ \qquad
Assia Benbihi$^2$ \qquad
Torsten Sattler$^2$ \qquad
Fredrik Kahl$^1$\\
$^1$Chalmers University of Technology\\
$^2$Czech Institute of Informatics, Robotics and Cybernetics, Czech Technical University in Prague\\
{\tt\small chelani@chalmers.se}
}

\twocolumn[{
\maketitle

\vspace{-1.2em}

\centering
\includegraphics[width=0.88\textwidth]{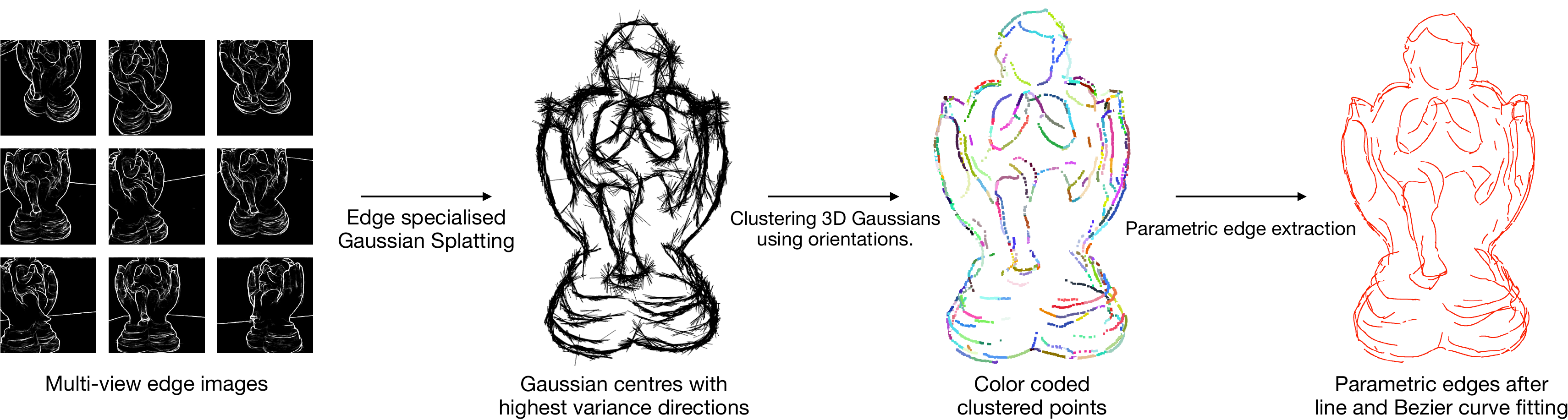}
\vspace{-0.25em}
\captionof{figure}{
\textbf{EdgeGaussians:} the proposed method learns 
     oriented 3D edge points via Gaussian Splatting specialized for 2D edge maps. The mean and the direction of largest variance of a Gaussian define an edge point's position and orientation. 
    Left to right: 2D edge maps generated by off-the-shelf detectors~\cite{su2021pixel,poma2020dense} are used as supervisory signals to train the 3D edge Gaussians. %
    The trained Gaussians are clustered based on spatial proximity and orientation consistency.
    Parametric edges are fitted on top of these clusters.
\label{fig:pipeline}
}
\vspace{1.5em}
}]

\begin{abstract}
With their meaningful geometry and omnipresence in the 3D world, edges are extremely useful primitives in computer vision. 
Methods for 3D edge reconstruction have 1) either focused on reconstructing 3D edges by triangulating tracks of 2D line segments across images or 2) more recently, learning a 3D edge distance field from multi-view images.
The triangulation-based methods struggle to repeatedly detect and robustly match line segments resulting in noisy and incomplete reconstructions in many cases.
Methods in the latter class rely on sampling edge points from the learnt implicit field, which is limited by the spatial resolution of the voxel grid used for sampling, resulting in imprecise points that require refinement. Further, such methods require a long training that scales poorly with the size of the scene. In this paper, we propose a method that explicitly learns 3D edge points with a 3D Gaussian Splatting representation trained from edge images. The 3D Gaussians are regularized to have their directions of largest variance along the edge they lie on, enabling clustering into separate edges. Backed by efficient training, the proposed method produces results better than or at-par with the current state-of-the-art methods, while being an order of magnitude faster. Code released at \href{https://github.com/kunalchelani/EdgeGaussians.}{https://github.com/kunalchelani/EdgeGaussians.}

\end{abstract}

\section{Introduction}

Edges are one of the main visual primitives that intelligent systems identify during visual processing~\cite{marr1980theory,hubel1962receptive,marr2010vision}.
They represent the boundaries of a scene, which are relevant for various computer vision tasks such as mapping~\cite{liu2023limap,ramalingam2015line,bartoli2005structure}, localization~\cite{liu1990determination,lee2019elaborate,hruby2024handbook}, place recognition~\cite{benbihi2020image,liu2019stochastic,taubner2020lcd},
surface reconstruction enhancement~\cite{alexa2003computing,huang2013edge,guennebaud2004real,oztireli2009feature}, visual odometry~\cite{kim2018edge,kneip2015sdicp,zhou2018canny,wu2020robust,wu2019semantic}, \gls{slam}~\cite{smith2006real,smith2006real}, and rendering~\cite{celes2011fast,tang2010stable,celes2010texture}.

3D edges comprise of straight 3D lines and 3D curves, which we will refer to respectively as lines and curves for simplicity.
Seminal works reconstruct 3D lines from images with a \gls{sfm} approach where line segments are detected from images, matched, and triangulated into 3D lines~\cite{weng1992motion,weng1988estimating,baillard1999automatic,montiel2000structure,bartoli2005structure,ramalingam2015line}.
The approach extends to 3D curves~\cite{kahl2003multiview,schmid2000geometry,robert1991curve,kaminski2001multiple}.
However, the lack of repeatability in the 2D edge detection and the limited robustness in the edge matching remain performance bottlenecks for such approaches. although recent works have made impressive progress in mapping lines~\cite{pautrat2021sold2,pautrat2023deeplsd,liu2023limap}.


3D edge extraction from 3D point clouds is free of those limitations and usually follows three steps: classify which 3D points lie on edges, cluster the points belonging to the same edge, and link them~\cite{weber2010sharp,bazazian2015fast,yu2018ec,wang2020pie}.
However, the classification is usually hindered by the extreme imbalance between the edge point and the non-edge points, and noisy point clouds can lead to spurious classification.
An efficient alternative is to operate directly on a 3D edge point cloud, \ie, a point cloud with points only on the 3D edges.

Recent methods~\cite{xue2024neat,Li2024CVPR,Ye_2023_CVPR} have proposed to sample such an \textit{edge point cloud} from neural implicit fields learned to represent 3D edges using multi-view images or 2D edge maps~\cite{dollar2014fast,su2021pixel} as supervision.
Parametric 3D edges are then fit on these sampled edge points.
However, such neural fields are computationally expensive and require long training times: recent methods\cite{Ye_2023_CVPR,xue2024neat,Li2024CVPR} take several hours to train on a simple CAD model from the \textit{ABC}~\cite{Ye_2023_CVPR,koch2019abc} dataset (see~\cref{sec:results} for precise runtimes).
Another limitation is the accuracy at which neural implicit fields can represent 3D edges: in theory, the 3D edge points lie on a level set of the field (0 for distance fields~\cite{Li2024CVPR,xue2024neat} and 1 for probability fields~\cite{Ye_2023_CVPR}).
In practice though, the field is evaluated at a finite 3D resolution and sampling points at the exact level set is not feasible.  
To compensate for this, the point sampling is done within an $\epsilon$-bound of the level sets~\cite{Ye_2023_CVPR,Li2024CVPR} but such points do not lie accurately on the 3D edges, requiring post-processing to correct these errors.

To make the 3D edge reconstruction simpler and more efficient while preserving accuracy,
we propose to learn an explicit representation of the 3D edge points.
Our method directly learns an edge point cloud, bypassing the need for level-set point sampling and post-processing.
It also directly learns the edge direction at each point, instead of requiring an additional step to infer it~\cite{Li2024CVPR}. 
The associated edge direction to each point makes the 3D edge fitting simpler.
Last, the method is several times faster (30 and 15 times as compared to ~\cite{Li2024CVPR} and ~\cite{Ye_2023_CVPR}, respectively), while producing results at-par with or slightly better than the current state-of-the-art.  
The proposed method optimizes 3D Gaussians to have their means close to the 3D edges and their direction of largest variance to be along the direction of the edge. 
These 3D Gaussians are then mapped to oriented points. 
Such a representation is geometrically meaningful and is trained in a fast and straightforward manner by adopting the training optimization defined in 3D Gaussian Splatting (3DGS)~\cite{kerbl3Dgaussians}.
The optimization is adapted to accommodate the specificities of 3D edge learning mainly the sparsity and the occlusion of the 3D edges.
The training remains simple and is supervised with off-the-shelf 2D edge maps~\cite{su2021pixel,poma2020dense}, as in~\cite{Li2024CVPR,Ye_2023_CVPR}.

To summarize, we make the following contributions:
i) We propose a simple, accurate, and extremely efficient method to reconstruct 3D edge points.
ii) The proposed method directly learns the oriented 3D points that form the 3D edge point cloud, hence bypassing the delicate and noisy level-set sampling of 3D edge points in existing implicit formulations.
iii) Results show that the proposed representation enables 3D edge reconstruction performance better than or at par with previous learning-based methods while running an order of magnitude faster.

\section{Related work}

We first review the 3D edge reconstruction methods based on multi-view images - the seminal \gls{sfm}-based methods and the more recent ones that learn implicit neural edge fields. We then discuss the methods for extracting 3D edges from point clouds.We also review 3D Gaussian Splatting methods, especially those with geometry constraints.

\vspace{0.2cm}
\PAR{3D Edges from multi-view images.}
Traditional methods reconstruct 3D lines from images with an \gls{sfm} approach
that detects lines~\cite{von2008lsd,pautrat2023deeplsd,huang2020tp,xu2021line,abdellali2021l2d2}, matches them across images based on line descriptors~\cite{pautrat2021sold2,lange2019dld,verhagen2014scale,bay2005wide}, and lifts them to 3D with triangulation~\cite{weng1988estimating,weng1992motion,zhang1995estimating,baillard1999automatic,montiel2000structure,bartoli2005structure,chandraker2009moving,zhang2014structure,micusik2017structure,liu2023limap,ramalingam2015line,do1961machine} or epipolar geometry~\cite{hofer2014improving,hofer2015line3d,hofer2017efficient}.
The main challenges are repeatedly detecting lines even if they are occluded and matching lines robustly across images.
While recent works~\cite{liu2023limap,hofer2014improving,hofer2015line3d,hofer2017efficient,hofer2015line3dpp} have made impressive progress in addressing these limitations, the robustness of the line detection and matching remains a performance bottleneck.
These limitations also hold for similar methods that reconstruct 3D curves~\cite{kahl2003multiview,schmid2000geometry,robert1991curve,kaminski2001multiple}.
Alternative approaches work around line matching by estimating 3D lines using geometric graph optimization on the detected 2D line segments~\cite{jain2010exploiting,ramalingam2013lifting}, or directly predicting the 3D wireframe in an end-to-end learned manner~\cite{zhou2019learning,ma20223d} but they fall behind \gls{sfm}-based method in reconstruction accuracy~\cite{liu2023limap}.
In this paper, we propose a method that is on-par with the state-of-the-art without relying on edge detection and matching.

Recent works~\cite{Ye_2023_CVPR,xue2024neat,Li2024CVPR} avoid the limitations of \gls{sfm}-based methods
by learning a neural implicit field to represent the 3D edges. 
The supervision for training such a field are natural RGB images~\cite{xue2024neat} or 2D edge maps~\cite{Ye_2023_CVPR,Li2024CVPR} generated by edge detectors~\cite{dollar2014fast,su2021pixel,poma2020dense,xue2023holistically}.
The resulting 3D field is then sampled to get the 3D edge points on top of which point clustering, linkage, and curve fitting are performed.
NEF~\cite{Ye_2023_CVPR} learns an edge density field that represents the probability of a 3D point to lie on a 3D edge.
The point sampling then amounts to sampling the 1-level-set.
Inspired by VolSDF~\cite{yariv2021volume}, the edge density is tied to a volumetric density~\cite{mildenhall2021nerf} so that NEF can be trained simply with volumetric rendering on the 2D edge maps~\cite{su2021pixel}.
EMAP~\cite{Li2024CVPR} improves on NEF~\cite{Ye_2023_CVPR} and produces state-of-the-art results by learning a 3D edge field with an \gls{udf}.
Both methods account for the class imbalance and the view-inconsistent occlusions of 2D edge maps with weighted rendering loss~\cite{Ye_2023_CVPR} and importance sampling of rays ~\cite{Li2024CVPR}.
NEAT~\cite{xue2024neat} adopts a \gls{sdf}~\cite{yariv2021volume}
to learn a 3D wireframe field from images jointly with junction points to derive the wireframes.
One common limitation of these methods is their reliance on computationally heavy neural implicit representations resulting in long training times.
Also, the edge points are sampled over level sets approximated with voxel grids so, even with a fine resolution, such points do not lie accurately on the 3D edges and require further refinement.
Instead, our method explicitly learns oriented points along the 3D edges in the form of 3D Gaussians.
These oriented points are then clustered into individual edges which are then represented in parametric form. 

\vspace{0.2cm}
\PAR{3D Edge extraction from point clouds.}
To extract edges from a set of 3D points, most approaches first classify which 3D points lie on edges, cluster the points that lie on the same edge, link them, and fit a parametric edge to each cluster.
Each step has hand-crafted~\cite{weber2010sharp,bazazian2015fast,yang2014automated,ahmed2018edge,hackel2016contour,demarsin2007detection} and learned variants~\cite{yu2018ec,wang2020pie,liu2021pc2wf,himeur2021pcednet,zhu2023nerve} and some methods operate in an end-to-end manner~\cite{wang2020pie,liu2021pc2wf,cherenkova2023sepicnet}. Such methods generally require a dense and noise-free point cloud as input and do not directly work on \gls{sfm} point clouds obtained from multi-view images.

\vspace{0.2cm}
\PAR{3D Gaussian Splatting.} The seminal work of~\cite{kerbl3Dgaussians} introduces the representation of scenes as
3D Gaussians which can be rendered efficiently.
The parameters of Gaussians are learned using a loss function that compares the rendered and ground-truth images from multiple views.
Similar to our approach, certain works introduced geometric regularization in 2D or 3D to model human hair strands using 3D Gaussians~\cite{zakharov2024gh,luo2024gaussianhair}.
However, they have been applied and tested specifically for the task of hair modeling.

\section{3D Edge Reconstruction with 3D Gaussians} %
\label{sec:method}



\begin{figure}
    \centering
    \includegraphics[width=0.8\linewidth]{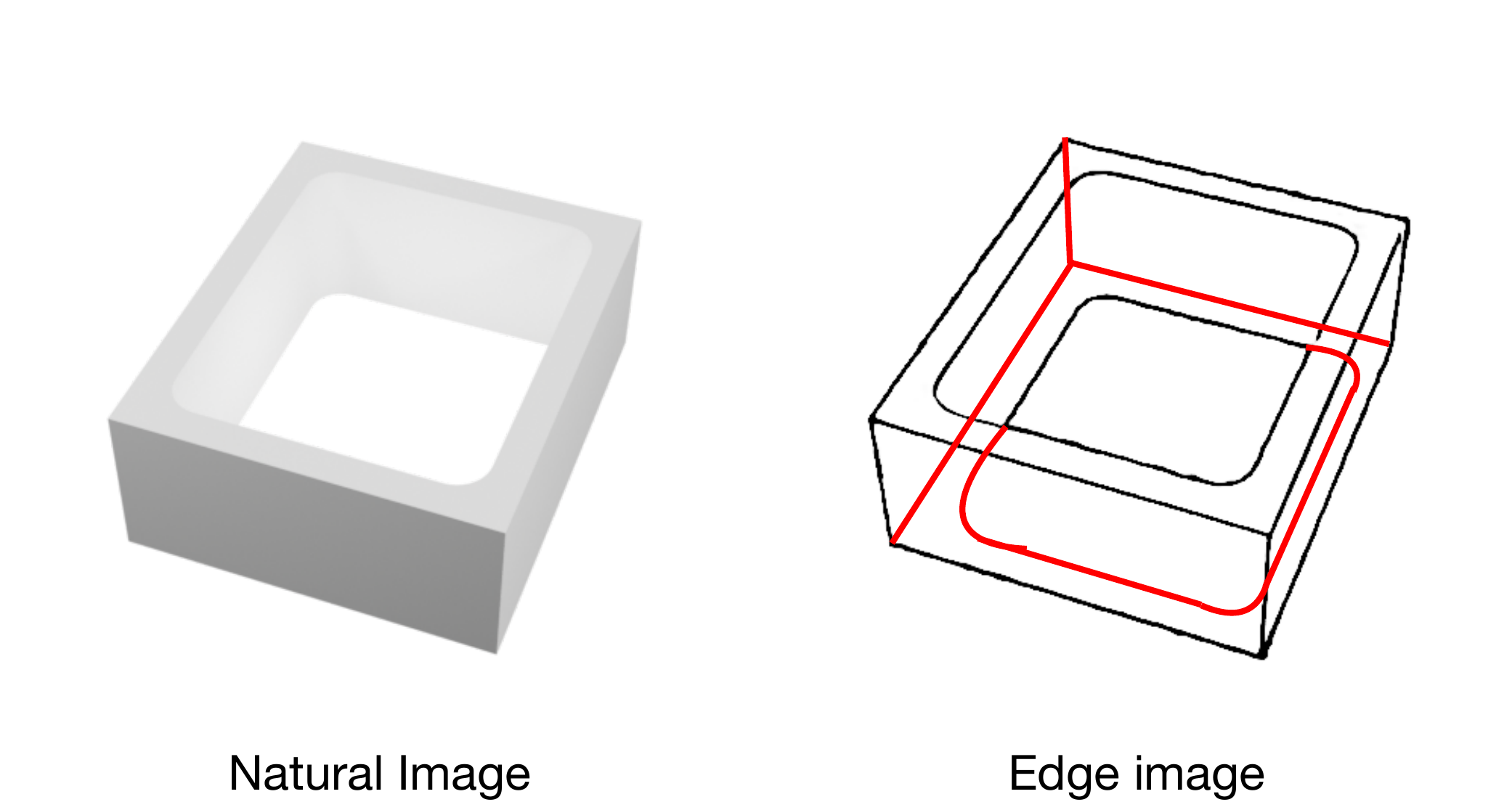}
    \caption{
    \textbf{Occlusions for 3D edges.}
    The red edges are the edges occluded by surface and are absent from the supervisory 2D edge maps. Yet, these edges are present in the rendering of the 3D edge representation, which is the desired behavior.
    }
    \label{fig:occlusion_explain}
\end{figure}

Given a set of edge images, our method directly derives oriented points along 3D edges. These points are then clustered into individual edges, on which parametric fitting is performed. 
A relevant learnable representation of an oriented 3D edge point is a 3D Gaussian; with the mean of the Gaussian being the point's position and its principal direction being the point's orientation.
Such a representation is efficiently trained using the 3DGS~\cite{kerbl3Dgaussians} optimization with edge images as input.
In this section, we first review the original 3DGS~\cite{kerbl3Dgaussians} framework. 
We then discuss how 3DGS is adapted for 3D edge reconstruction from multi-view 2D edge maps,
and lastly, we describe the parametric edge fitting from the optimized 3D Gaussians.

\subsection{Preliminaries: 3D Gaussian Splatting (3DGS)}
3D Gaussian Splatting~\cite{kerbl3Dgaussians} represents the scene using a set of 3D Gaussians.
A 3D Gaussian centered at $\mu$, with covariance $\Sigma$ is defined as:
\begin{equation}
    G(x) = e^{-\frac{1}{2}(x-\mu)^{T}\Sigma^{-1}(x-\mu)} \enspace . 
\end{equation}
Each Gaussian also has an opacity attribute $\alpha$ and a color attribute $c$ represented by spherical harmonic coefficients.
All the Gaussian's parameters are differentiable and are trained by rendering the 3D Gaussians and comparing the rendered images to the original images.
The loss is the sum of the $\mathcal{L}_1$ loss and the Difference of Structural Similarity loss~\cite{wang2004image} (D-SSIM) $\mathcal{L}_{\text{SSIM}}$ between the two, weighted by $\lambda \in \mathbb{R}$:
\begin{equation} \label{eq:gs_loss}
    \mathcal{L} = \lambda\mathcal{L}_{1} + (1 - \lambda)\mathcal{L}_{\text{D-SSIM}} \enspace .
\end{equation}
The rendering first projects the 3D Gaussians to 2D splats~\cite{zwicker2001ewa}, and the color of a pixel is then derived by $\alpha$-blending the colors of the Gaussians projecting on the pixel. 
To optimize the covariance matrix $\Sigma$ while maintaining its positive-semi definiteness, it is decomposed into a rotation matrix $R \in \mathbb{R}^{3 \times 3}$ and a diagonal scaling matrix $S \in \mathbb{R}^{3 \times 3}$, such that $\Sigma = RSS^{T}R^{T}$.

The representation is initialized with Gaussians centered at a sparse set of points, \eg, obtained from \gls{sfm}~\cite{schonberger2016structure}.
The set of Gaussians in the scene is dynamically controlled by duplicating, splitting, and culling Gaussians based on various criteria - the important ones being 1) culling low-opacity Gaussians; and 2) duplicating/splitting Gaussians that are in a region that needs densification, or a complex 3D surface requiring more Gaussians.

\subsection{Gaussian Splatting for Edge Images}
\label{subsec:gs_edge}

3DGS~\cite{kerbl3Dgaussians} is designed for fast and accurate novel view synthesis while our focus here is more on positioning of the Gaussians along 3D edges.
Additionally, the supervision in our method comes 2D edge maps instead of natural RGB images.
These differences in the final objective and the  data modality motivate certain changes in the training paradigm.
We first detail the issues that arise when using edge images for supervision and propose a solution to address these issues. 
Then we present a geometric regularization to steer the training objective from purely view synthesis, towards one focused on the geometry of 3D Gaussians useful for edge modeling.   

\begin{figure}
    \centering
    \includegraphics[width=0.8\linewidth]{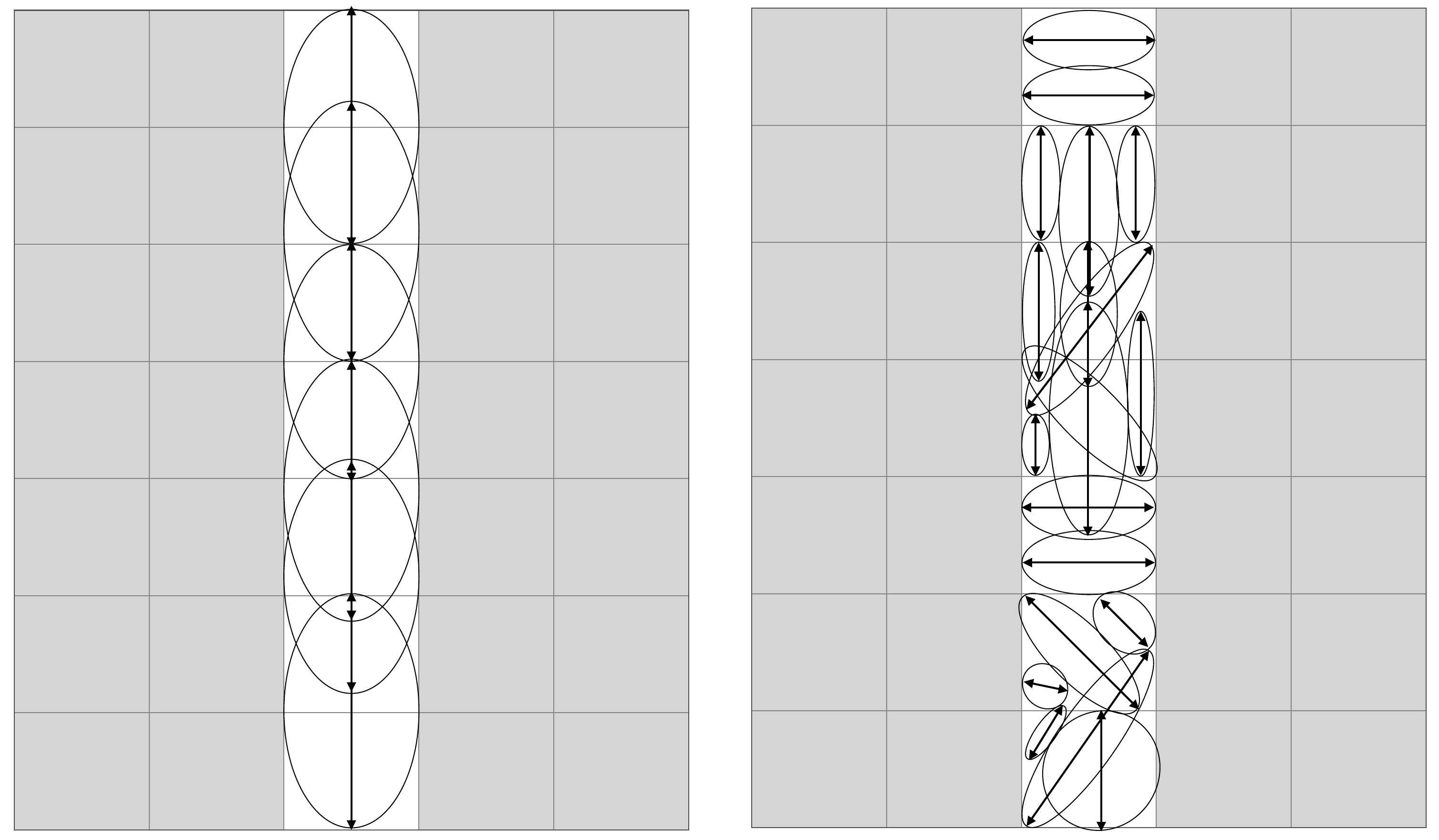}
    \caption{\textbf{Geometric regularization of 3D Gaussians.} Enforcing elliptical Gaussians with the principal direction aligned with the neighbors would result in 2D projections as shown on the left, while un-regularized ones might look like those on the right. The ones on the left are geometrically meaningful and result in easier clustering into separate edges.}
    \label{fig:orient_explain}
\end{figure}

\vspace{0.2cm}
\PAR{Supervision from edge maps.}
Natural images typically depict the surfaces of the scene, which is a dense 3D geometry.
On the other hand, 2D edge maps depict edges that are inherently sparse.
As already observed in previous works~\cite{Li2024CVPR,Ye_2023_CVPR}, such an extreme imbalance in the pixel distribution can hinder the training stability.
For example, since edge maps consist mostly of zero-value pixels where no 3D edge reprojects and few pixels with non-zero values, the training could converge to valid but degenerate configurations such as an empty set or a set Gaussians with only background color.

Additionally, 2D edge maps~\cite{dollar2014fast,su2021pixel,poma2020dense,xue2023holistically} exhibit occlusions due to the scene's surfaces that do not occur when rendering 3D edges. This makes the supervision signal incomplete as illustrated in~\cref{fig:occlusion_explain}: the supervisory edge map is generated from the natural image (left) and comprises only the visible edges (black edges on the right).
However, the rendering of the 3D edges learned by the trained representations do show the occluded edges (red), which is the desired behavior.
Naively training the model with the standard rendering loss would penalize the model for rendering the occluded edges, \ie, the red edges.

To address this issue, previous works~\cite{Ye_2023_CVPR,Li2024CVPR} 
weight the rendering loss higher 
for edge pixels as compared to non-edge pixels~\cite{Ye_2023_CVPR}, or compute the loss over pixels sampled uniformly across the edge and non-edge pixels~\cite{Li2024CVPR}.
In our experiments, 
we observe that the former is more efficient while the latter leads to slightly better results, which we adopt and detail next.

The sampling is derived by masking the $\mathcal{L}_1$ loss between a rendered image $\hat{I}$ and a training image $I$ as follows:
\begin{equation}
	\mathcal{L}_{\text{proj}} = \text{average} \left (\mathcal{M} \odot |\hat{I} - I|\right) \enspace
\end{equation}
where $\odot$ is the Hadamard product operation and $\mathcal{M}$ is a 2D binary mask with values 1 at all edge pixels and at an equal number of randomly selected background pixels.
Also, we discard the SSIM term in~\cref{eq:gs_loss}.

\vspace{0.2cm}
\PAR{Learning oriented edge points as Gaussians.}
The 3D Gaussians optimized only with the rendering loss can be oriented and scaled 
without correlation to the 3D edge geometry.
This impedes the subsequent edge fitting so we introduce geometric regularization 
illustrated in~\cref{fig:orient_explain}.
As later detailed in~\cref{subsec:edge_fit}, we adopt a cluster-then-fit strategy that first
clusters the points into groups that lie on the same edge and then fits an edge to each cluster.

Clustering edges only based on spatial proximity 
raises ambiguities when clustering points close to two or more edges - points close to junctions.
A standard way to address such ambiguity is to cluster points that are not only spatially close but also have the same orientation~\cite{ni2016edge}.
To enable this, the learned edge points must have the same orientation as the edge they lie on.
This is equivalent to enforcing the \textit{principal direction of each 3D Gaussian}, \ie, the direction of its largest variance, to align with the edge's direction.
In practice, we implement such a constraint by encouraging the principal direction of a 3D Gaussian to point towards its $k$ nearest neighbors in 3D, $k$ being a hyperparameter.

Given $N$ 3D Gaussians indexed between $1$ and $N$, let $\mu_{i}$ and $d_i$
be respectively the center and the principal direction
of the $i^{th}$ Gaussian,
and let $i_1, ..., i_k$ be the indices of the neighbors,
we enforce the constraint with the following loss:
\begin{equation}\label{eq:orient_loss}
	\mathcal{L}_{\text{orient}} = 1 - \frac{1}{N}\left(\sum_{i=1}^{N}\frac{1}{k}\sum_{j=1}^{k}|d_{i}^{T} d_{i_{j}}|\right)
\end{equation}
Note that this approach works in general when the $k$ nearest neighbors lie on the same edge and are optimized to approximately correct positions.
However, it can 
break down in the presence of fine structures as shown in the failure cases.

We also regularize the Gaussians to have an ellipsoidal shape instead of a spherical or disc-like shape so that identifying the principal direction of the Gaussian, \ie, the largest axis, is computationally more stable.
This amounts to enforcing the ratio between the largest scale and the next one to be large:
\begin{equation}\label{eq:shape_loss}
	\mathcal{L}_{\text{shape}} = \frac{1}{N}\sum_{i=1}^{N}\frac{{}^{2}s_{i}}{{}^{1}s_{i}}
\end{equation}
where ${}^{1}s_{i}$ and ${}^{2}s_{i}$ are the scales of the largest and the second largest axes of the $i^{th}$ Gaussian.

The final loss is the sum of the three terms weighted by $\lambda_{1}, \lambda_{2} \in \mathbb{R}$:
\begin{equation}\label{eq:total_loss}
	\mathcal{L} = \mathcal{L}_{\text{proj}} + \lambda_{1}\mathcal{L}_{\text{orient}} + \lambda_{2}\mathcal{L}_{\text{shape}}
\end{equation}
An example of the resulting 3D Gaussians is shown in~\cref{fig:pipeline} where the length of the Gaussian's principal direction is increased for visualization purposes.

\subsection{Edge Fitting}
\label{subsec:edge_fit}

Given a 3D edge point cloud, there are mainly two edge fitting strategies: the fit-then-cluster approach that operates in a multi-RANSAC~\cite{fischler1981random} fashion.
The clusters are simply the inlier supports of each edge.
The second one is the cluster-then-fit approach in which points are first clustered into individual edges.
The first one is used in NEF~\cite{Ye_2023_CVPR} while the second one is used in EMAP~\cite{Li2024CVPR}.
The fit-then-cluster approach requires more engineering to prevent points from incorrectly supporting an edge and disambiguation between fitting a line or a low-curvature edge.
We therefore adopt the simpler cluster-then-fit strategy which is detailed next.

\PAR{Clustering points into edges.} Graph traversal is a common strategy to cluster points 
where points form the vertices of the graph and an edge exists between two vertices only if they are spatially close~\cite{mineo2019novel,grant1981efficient}.
Clustering the points into their supporting edges then amounts to solving a graph traversal problem.
This simple solution is usually made robust by integrating a smoothness constraint between neighboring points~\cite{ni2016edge}: two vertices are connected not only if they are spatially close but if their direction is also similar.
This strategy is used in~\cite{Li2024CVPR} where the point's direction is derived after training the edge field with a subsequent optimization whereas our method is simpler in that it learns both the edge point position and orientation directly.

Given the trained Gaussians, we define the graph as follows.
The vertices are the oriented edge points: the vertex's position is the Gaussian's center and the vertex's direction is the Gaussian's principal direction.
Vertices are neighbors if they are spatial neighbors and if they have similar orientations.
A cluster is initialized with a point and a new point is added to the cluster if it is a neighbor of the last added point and if its orientation aligns with the orientation of the edge grown so far.
To avoid adding points that do not lie on the edge but are close and oriented parallel to the edge, we also check if the new point's orientation aligns with the line joining the last added point and the new one.
For both orientation tests, we use a single orientation threshold $\theta$.
The graph traversal results in clusters of points belonging to the same edge.
We then fit edges to the resulting clusters in the form of line segments and cubic B\'ezier curves.

\begin{table*}[t]
\centering
\resizebox{0.85\textwidth}{!}{ %
\begin{tabular}{l|l|l|c c|c c c|c c c|c c c}
\toprule
\textbf{Method} & \textbf{Detector} & \textbf{Modal} & \textbf{Acc$\downarrow$} & \textbf{Comp$\downarrow$} & \textbf{R5$\uparrow$} & \textbf{R10$\uparrow$} & \textbf{R20$\uparrow$} & \textbf{P5$\uparrow$} & \textbf{P10$\uparrow$} & \textbf{P20$\uparrow$} & \textbf{F5$\uparrow$} & \textbf{F10$\uparrow$} & \textbf{F20$\uparrow$} \\
\midrule
LIMAP~\cite{liu2023limap}  & LSD & Line & 9.9 & 18.7 & 36.2 & 82.3 & 87.9 & 43.0 & 87.6 & 93.9 & 39.0 & 84.3 & 90.4 \\
 & SOLD2 & Line & \textbf{5.9} & 29.6 & \textbf{64.2} & 76.6 & 79.6 & \textbf{88.1} & \textbf{96.4} & \textbf{97.9} & \textbf{72.9} & 84.0 & 86.7 \\
 \midrule
 & PiDiNeT$^{\dagger}$ & Curve & 11.9 & 16.9 & 11.4 & 62.0 & 91.3 & 15.7 & 68.5 & 96.3 & 13.0 & 64.0 & 93.3 \\
NEF~\cite{Ye_2023_CVPR}  & PiDiNeT & Curve & 15.1 & 16.5 & 11.7 & 53.3 & 93.9 & 12.3 & 61.3 & 95.8 & 12.3 & 51.8 & 88.7 \\
 & DexiNed & Curve & 21.9 & 15.7 & 11.3 & 48.3 & 93.7 & 11.5 & 58.9 & 91.7 & 10.8 & 42.1 & 76.8 \\
 \midrule
 & PiDiNeT & Edge & 9.2 & 15.6 & 30.2 & 75.7 & 89.5 & 35.6 & 79.1 & 95.4 & 32.4 & 77.0 & 92.2 \\
EMAP~\cite{Li2024CVPR} & DexiNed & Edge & 8.8 & 8.9 & 56.4 & 88.9 & 94.8 & 62.9 & 89.9 & 95.7 & 59.1 & 88.9 & 94.9 \\
\midrule
    & PiDiNeT & Edge & 11.7 & 10.3 & 17.1 & 73.9 & 83.1 & 26.0 & 87.2 & 92.5 & 20.6 & 79.3 & 86.7 \\
Ours & DexiNed & Edge & 9.6 & \textbf{8.4} & 42.4 & \textbf{91.7} & \textbf{95.8} & 49.1 & 94.8 & 96.3 & 45.2 & \textbf{93.7} & \textbf{95.7} \\ 
\bottomrule
\end{tabular}
}
\caption{\textbf{3D Edge Reconstruction on \textit{ABC-NEF}~\cite{Ye_2023_CVPR}.}
Overall, the proposed method is on par with the baselines and slightly better than the implicit representations NEF~\cite{Ye_2023_CVPR} and EMAP~\cite{Li2024CVPR} under the 10mm and 20mm error thresholds.
Under the 5mm threshold, the geometry-based LIMAP~\cite{liu2023limap} outperforms most methods.
We explain the lower performance of our method under this threshold by the bias in the 2D edge maps~\cite{su2021pixel,poma2020dense} used as a supervisory signal.
The baselines results are taken as provided by Table 1 in EMAP~\cite{Li2024CVPR}.
}
\label{tab:abc_quant_comp}
\end{table*}



\PAR{Parametric Edge Fitting.}
Once the points are clustered into individual edges, we next select whether a line or a cubic B\'ezier curve best fits the set of points.
One solution~\cite{Li2024CVPR} consists is first fitting lines until no more lines can be fitted with a geometric error below a given threshold then fitting curves to the remaining points.
However, with edges of different lengths and different noise levels, tuning the geometric error threshold is complex and can result in many curves incorrectly modeled as lines, as observed in the EMAP~\cite{Li2024CVPR} results.
Instead, we propose to compare both the line and the curve model for each point cluster and select the best based on their relative geometric errors.

\begin{figure*}[t]
    \centering
    \includegraphics[width=0.90\linewidth]{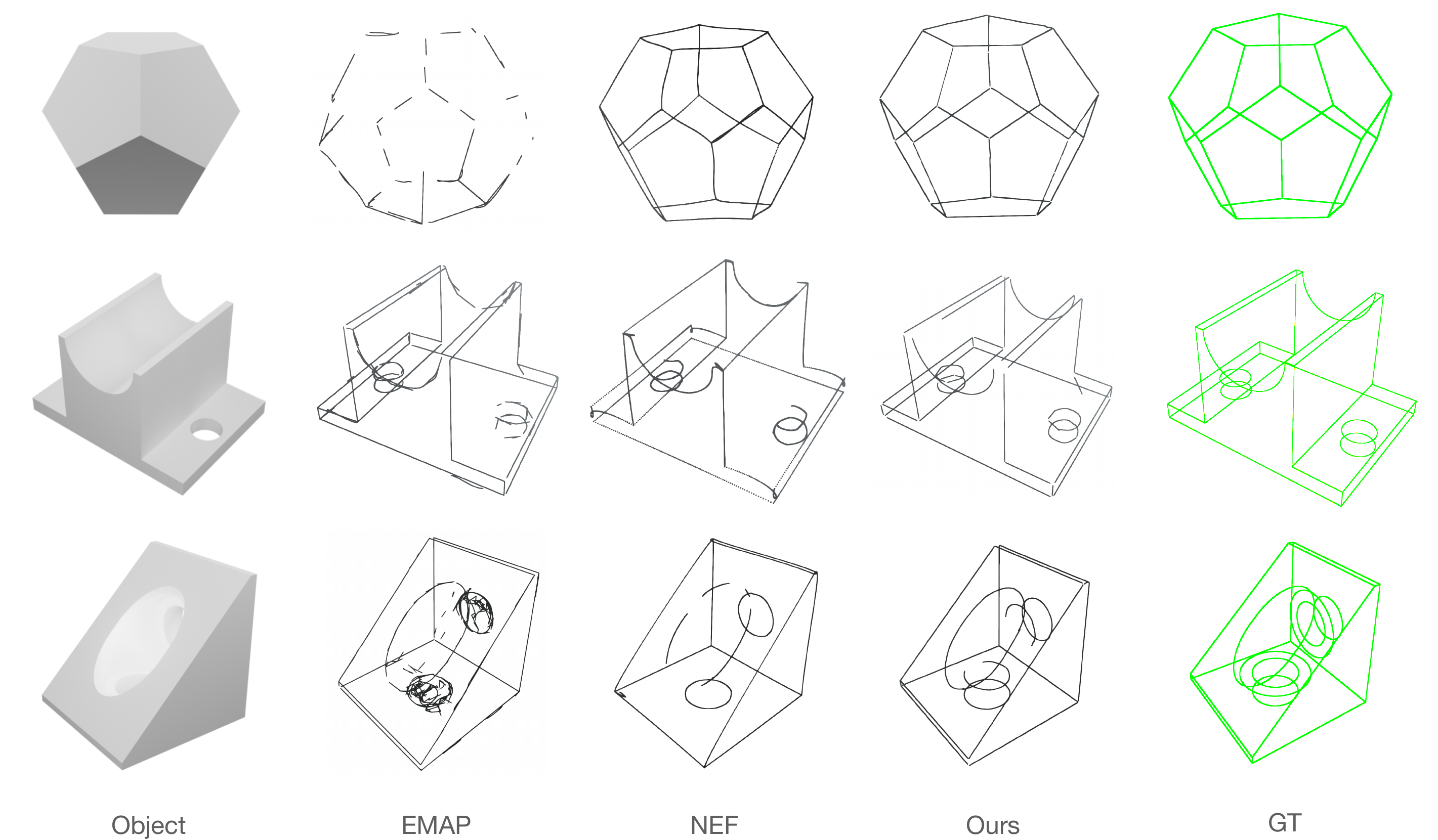}
    \caption{\textbf{Qualitative results on \textit{ABC-NEF}~\cite{Ye_2023_CVPR}.}
    The proposed method captures curves and lines accurately but can be marginally incomplete.
    EMAP~\cite{Li2024CVPR} is either slightly less complete than the proposed method or predicts extra edges.
    NEF~\cite{Ye_2023_CVPR} produces correct edges 
    but exhibits knots observed around corners on many occasions.}
    \label{fig:abc_quali}
\end{figure*}

Let $e_{\text{c}}$ and $e_{\textit{l}}$ be the average residual error of the curve and line models respectively.
We select the curve model if the error $e_{\text{c}}$ is smaller than a fraction $\delta$ of the line residual error, \ie, $e_{\text{c}} \leq \delta e_{\textit{l}}$.
The parameter $\delta$ can be tuned to control the fraction of lines and curves in the output.
As shown in the qualitative results (\cref{fig:abc_quali}), this strategy is more effective than the ones adopted in NEF~\cite{Ye_2023_CVPR} and EMAP~\cite{Li2024CVPR}.
NEF~\cite{Ye_2023_CVPR} first approximates all edges with 2-control-points Bezier curves, \ie, 3D lines, and uses the estimated control point to further optimize the models into 3D curves when needed.
This approach produces several degenerate curve configurations, especially close to the corners.

\section{Evaluation}

We evaluate the proposed method against image-based edge reconstruction methods following the evaluation setup in EMAP~\cite{Li2024CVPR}.
Results show that our method reaches mapping performance on par with the state-of-art while running an order of magnitude faster.
The rest of this section describes the evaluation setup and reports the results.

\label{sec:results}

\subsection{Experimental Setup}
\label{subsec:xp_setup}

\PAR{Datasets.}
The quantitative evaluation is performed on the subsets of \textit{ABC-NEF}~\cite{Ye_2023_CVPR} and the DTU~\cite{jensen2014large} datasets used by EMAP~\cite{Li2024CVPR} for evaluation.
For fairness, we train on the 2D edge maps ~\cite{su2021pixel,poma2020dense} they provide and use their released evaluation code.
Qualitative results on selected scenes from Replica~\cite{replica19arxiv} and TnT~\cite{Knapitsch2017} datasets are provided in the supplementary material.

The \textit{ABC-NEF~\cite{Ye_2023_CVPR}} subset of \textit{ABC} dataset~\cite{koch2019abc} is curated to evaluate 3D edge mapping. 
It consists of CAD models, ground truth parametric edges and 50 views rendered from around the object.
In EMAP~\cite{Li2024CVPR}, models with cylinders and spheres are removed as they are inconsistent shapes for the task of edge mapping.
Similarly, we use the 6 scenes from the DTU~\cite{jensen2014large} dataset that contains multi-view images of everyday objects captured from fixed frontal viewpoints.
A pseudo-ground truth of \textit{“edge-points”} is created by projecting the dense 3D points reconstructed by a structured-light scanner~\cite{aanaes2016large} onto the 2D edge maps.
Although we perform at par or better than the baselines on DTU, we observe that the pseudo-ground-truth introduces biases in the evaluation and we illustrate them in the qualitative results.
Finally, qualitative results on selected scenes from Replica~\cite{replica19arxiv} and TnT~\cite{Knapitsch2017} datasets are provided in the supplementary material.

\begin{table*}[t]
\centering
 \begin{subtable}[b]{0.7\linewidth}
 \fontsize{9}{6.5}\selectfont
    \centering
\begin{tabular}{l|c c|c c|c c|c c|c c}
\toprule
\textbf{Scan} & \multicolumn{2}{c|}{{LIMAP~\cite{liu2023limap}}} & \multicolumn{2}{c|}{{NEF~\cite{Ye_2023_CVPR}}} & \multicolumn{2}{c|}{{NEAT~\cite{xue2024neat}}} & \multicolumn{2}{c|}{{EMAP~\cite{Li2024CVPR}}} & \multicolumn{2}{c}{{Ours}} \\
 & \textbf{R5$\uparrow$} & \textbf{P5$\uparrow$} & \textbf{R5$\uparrow$} & \textbf{P5$\uparrow$} & \textbf{R5$\uparrow$} & \textbf{P5$\uparrow$} & \textbf{R5$\uparrow$} & \textbf{P5$\uparrow$} & \textbf{R5$\uparrow$} & \textbf{P5$\uparrow$} \\
\midrule
{37}  & 75.8 & 74.3 & 39.5 & 51.0 & 63.9 & 85.1 & 62.7 & 83.9 & \textbf{84.8} & \textbf{87.1} \\
{83}  & 75.7 & 50.7 & 32.0 & 21.8 & 72.3 & 52.4 & 72.3 & 61.5 & \textbf{86.4} & \textbf{64.8} \\
{105} & 79.1 & 64.9 & 30.3 & 32.0 & 68.9 & 73.3 & 78.5 & \textbf{78.0} & \textbf{81.7} & 76.8 \\
{110} & 79.7 & 65.3 & 31.2 & 40.2 & 64.3 & \textbf{79.6} & 90.9 & 68.3 & \textbf{92.9} & 57.2 \\
{118} & 59.4 & 62.0 & 15.3 & 25.2 & 59.0 & 71.1 & 75.3 & \textbf{78.1} & \textbf{86.0} & 77.6 \\
{122} & 79.9 & 79.2 & 15.1 & 29.1 & 70.0 & 82.0 & 85.3 & 82.9 & \textbf{94.8} & \textbf{86.9} \\
\midrule
{Mean} & 74.9 & 66.1 & 27.2 & 33.2 & 66.4 & 73.9 & 77.5 & \textbf{75.4} & \textbf{87.7} & 75.1 \\
\bottomrule
\end{tabular}
    \end{subtable}%
 \begin{subtable}[b]{0.28\linewidth}
 \fontsize{7}{6.2}\selectfont
        \centering
\begin{tabular}{l*{2}{c}}
\toprule
    & \textit{ABC-NEF}~\cite{Ye_2023_CVPR}       & DTU~\cite{jensen2014large} \\
\midrule
NEF~\cite{Ye_2023_CVPR}                 & 1:26   & 1:50 \\
NEAT~\cite{xue2024neat}                 & 14:13  & 8:38 \\
EMAP~\cite{Li2024CVPR}                  & 2:30   & 12:00 \\
\textbf{Ours}                           & 0:05   &  0:05    \\
\bottomrule
\end{tabular}
    \end{subtable}
    \caption{\textbf{Left: 3D Edge Reconstruction on DTU~\cite{jensen2014large}.}
Comparison of precision (P) and recall (R) under 5mm error.
The scenes are scaled to a bounding box of a maximum side of 1 meter.
\textbf{Right: Average Runtimes} of implicit representation methods in `hour:minutes'.
The 3D edge reconstruction of the proposed method performs better or is at par with the learning-based baselines NEF~\cite{Ye_2023_CVPR}, EMAP~\cite{Li2024CVPR} and NEAT~\cite{xue2024neat} and runs an order of magnitude faster.}
    \label{tab:dtu_and_runtime}
\end{table*}

\begin{figure*}[t]
    \centering
    \includegraphics[width=0.86\linewidth]{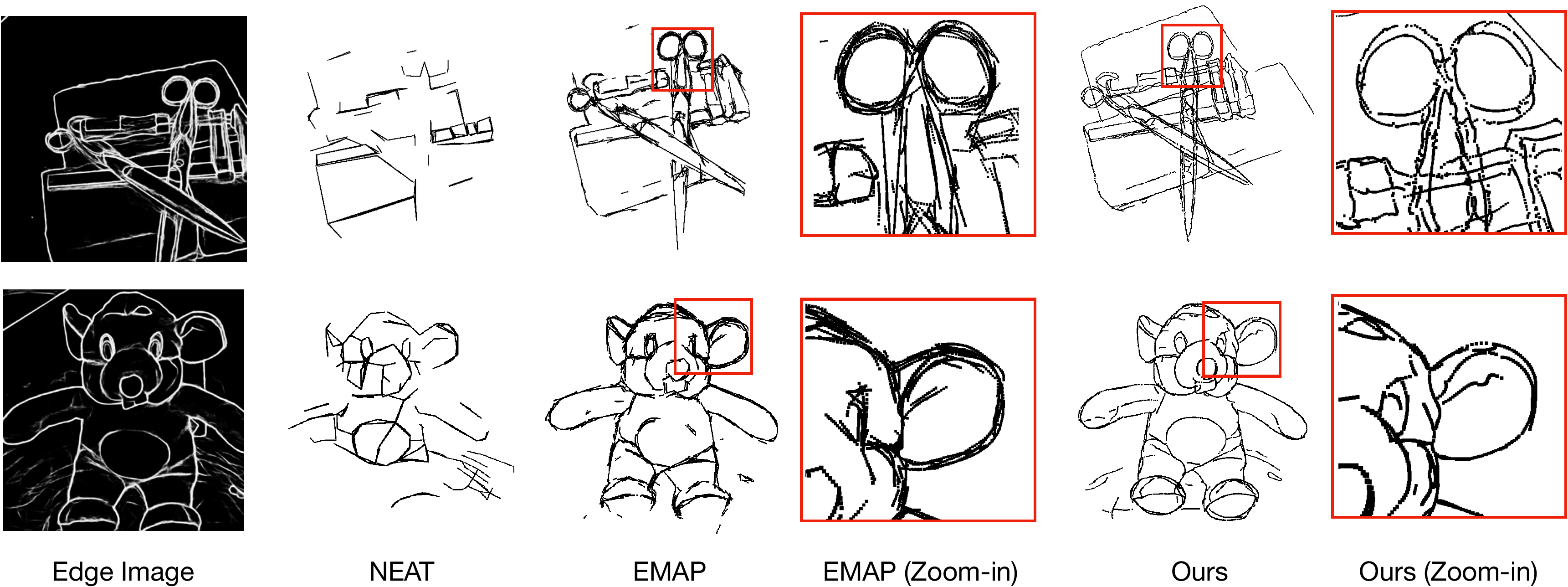}
    \caption{
    \textbf{Qualitative results on DTU~\cite{jensen2014large}.}
    Comparison of the proposed 3D reconstruction method 
    against baselines. 
    NEAT~\cite{xue2024neat} produces partially complete reconstructions with lines only.
    The edges from EMAP~\cite{Li2024CVPR} are more complete but there are duplicate edge predictions for a single target edge in 3D, while our reconstruction is much cleaner with mostly a single predicted edge boundary (see zoom-ins).}
    \label{fig:dtu_quant}
\end{figure*}

\PAR{Baselines.}
We evaluate the line-\gls{sfm}-based method LIMAP~\cite{liu2023limap} that sets the state-of-the-art in geometry-based 3D line reconstruction
and the 3D wireframe learning method NEAT~\cite{xue2024neat}.
We also evaluate the state-of-the-art 3D line / curve mapping methods
NEF~\cite{Ye_2023_CVPR} and EMAP~\cite{Li2024CVPR}.
Following~\cite{Li2024CVPR}, NEAT~\cite{xue2024neat} is not evaluated
on \textit{ABC-NEF}~\cite{Ye_2023_CVPR}: the authors observe that it often fails to train on the textureless renderings of the CAD models.
The qualitative results we report are generated with the code and the weights released by the authors of the various baselines.
For EMAP~\cite{Li2024CVPR}, at the time of writing, running the released code with the recommended configurations did not yield the same results as the ones shown in the paper or the website.
We therefore report the results run on checkpoint versions graciously provided by the authors.
The quantitative results for the baselines are reported as is from~\cite{Li2024CVPR}.

\PAR{Metrics.}
For quantitative evaluation, the process defined in~\cite{Li2024CVPR} samples points along the predicted parametric edges and compares those points against points sampled at the same resolution on the ground-truth edges to compute the metrics.
The \textit{accuracy} defines the mean distance from the predicted points to the closest ground-truth points, and the \textit{completeness} defines the mean distance between the ground-truth points and their nearest predicted point.
For these two metrics, the lower the better.
The \textit{precision} at a distance threshold $\tau$ (P($\tau$)) measures the percentage of predicted points with at least one ground-truth point within distance $\tau$.
Symmetrically, the \textit{recall} (R($\tau$)) measures the percentage of ground-truth points for which a predicted point is within the distance threshold $\tau$. %
For precision and recall, the higher the better.
We report the metrics for $\tau$ in 5, 10, and 20 millimeters (mm).
We also report the training times of the neural implicit representations to show the advantage of the explicit representation adopted in this paper.

Although these metrics provide a meaningful evaluation of the reconstruction quality, they do not account for all the performance aspects of the 3D edge reconstruction.
For example, duplicate edge reconstructions that lie close to a ground-truth edge are not penalized.
This is observed in several reconstructions from LIMAP~\cite{liu2023limap} and EMAP~\cite{Li2024CVPR}. Further, this is especially important on the DTU dataset where pseudo-ground-truth 3D edge points do not lie exactly on the 3D edges but close to them
(See~\cref{fig:dtu_gt}).

\textbf{Implementation details} are provided in the suppl. mat.

\subsection{Results} %

\begin{figure*}[t]
    \centering
    \includegraphics[width=0.86\linewidth]{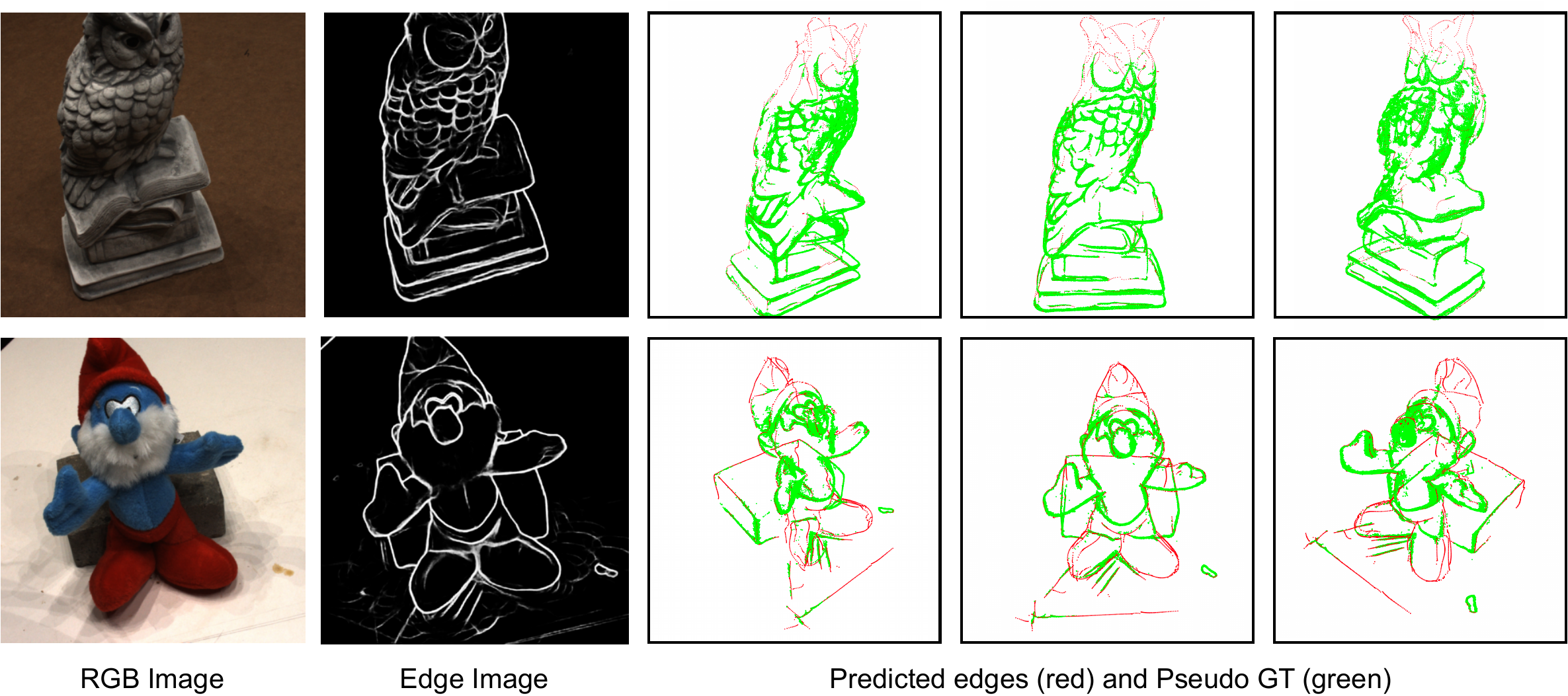}
    \caption{
    \textbf{Qualitative results and pseudo-ground-truth on DTU~\cite{jensen2014large}.}
    We visualize our method (red) and the pseudo-ground-truth edge points (green) generated in~\cite{Li2024CVPR} by projecting ground-truth 3D points onto edge images and using those projecting on edges in a majority of views~\cite{bignoli2018multi}.
    Note that our method gets penalized in precision when it predicts geometrically correct 3D edges that are missing from the pseudo-ground-truth, \eg, the top of the smurf and the howl.
    Also, the pseudo-ground-truth contains thick patches of points that make it difficult for well-predicted edges
    to get a high recall.
    Still, our method faithfully reconstructs the 3D parametric edges of the objects.}
    \label{fig:dtu_gt}
\end{figure*}

\PAR{Evaluation on \textit{ABC-NEF}~\cite{Ye_2023_CVPR}.}
We report the quantitative evaluation in~\cref{tab:abc_quant_comp} and the qualitative results in~\cref{fig:abc_quali}.
Overall, the proposed method is on par or slightly better than the implicit-representations-based NEF~\cite{Ye_2023_CVPR} and EMAP~\cite{Li2024CVPR}
and runs respectively 17 and 30 times faster (\cref{{tab:dtu_and_runtime}}-right).
This supports that the proposed explicit representation enables efficient 3D edge reconstruction while preserving accuracy.

Our method produces the most complete results while being almost as precise as the methods leading that field. LIMAP~\cite{liu2023limap} demonstrates impressive precision while missing curves and therefore lacking completeness.
EMAP~\cite{Li2024CVPR} performs better than our method under the 5mm error threshold although this trend is inverted under the 10mm and 20mm error thresholds.
The performance difference under the 5mm threshold can be attributed to a combination of two things. Firstly, EMAP performs an extra point refinement step, that our method does not. 
Secondly, there is a bias introduced by the thickness of 2D edge maps generated by the detectors~\cite{su2021pixel,poma2020dense}: 
we visually verify it by observing that the ground truth edges project close to the thick edges produced by these detectors, instead of projecting to its center.
Shifting or scaling the Gaussians appropriately to counter this bias is left for future work.

\PAR{Evaluation on DTU~\cite{jensen2014large}.}
The quantitative results are reported in~\cref{tab:dtu_and_runtime}-left and the comparative qualitative results are shown in~\cref{fig:dtu_quant}.
While our method either outperforms or is at par with the baselines as per the defined quantitative evaluation, we observe biases in the evaluation 
due to the pseudo-ground-truth generation process.
The pseudo-GT obtained by filtering out 3D points that do not project on 2D edge maps in enough views has two flaws. 
Firstly, the pseudo-GT points do not span the part of the scene that is not covered by a sufficient number of cameras. This issue is highlighted in ~\cref{fig:dtu_gt} where the pseudo-GT points in green show an incomplete coverage of the structure while our method correctly predicts a set of edges that completely cover the scene.
Secondly, the 3D points that get labeled as pseudo-ground-truth edge points are not only the points lying on the ground-truth edges but also the points within an $\epsilon$-bound on the edge, and the range of the bound is a function of the 2D edge map thickness and the distance between the object and the camera.
The first bias decreases the precision score for a set of edges that actually completely cover the scene as illustrated by our precision scores in~\cref{tab:dtu_and_runtime}.
As for the second bias, \ie, the bias of thick edges, it not only penalizes the recall but it also promotes the presence of duplicate edges, as estimated by EMAP~\cite{Ye_2023_CVPR}.
This can be seen in the examples shown in~\cref{fig:dtu_quant}.
Regarding the runtime, our method is as efficient as on the \textit{ABC-NEF}~\cite{Ye_2023_CVPR} dataset and runs 22 times faster than NEF~\cite{Li2024CVPR} and more than a hundred times faster than EMAP~\cite{Li2024CVPR} (\cref{tab:dtu_and_runtime}-right).

\section{Limitations and Future work}
The method inherits weaknesses of the original 3DGS~\cite{kerbl3Dgaussians} approach - it requires tuning of several parameters for the adaptive density control.
Also, while the geometric regularization and clustering work robustly for simple CAD objects, they sometimes lead to artifacts and inaccuracies in more complex scenes. A few such issues are presented in the supplementary material. Our clustering method is driven by heuristics and therefore occasionally fails in certain parts of the scene. Incorporating structural priors can make this component more robust. 

\section{Conclusion}
In this work, we propose a method for 3D edge reconstruction from images that explicitly learn the 3D edge points on top of which the 3D edges are fitted.
The explicit representation is simple and casts 3D edge points as 3D Gaussians and the edge direction as the principal axis of the Gaussians.
Such a representation allows for efficient rendering-based training
supervised with off-the-shelf 2D edge maps.
Results show that the proposed method is several times faster than the previous learning based approaches, while being slightly better or at-par in terms of the accuracy and completeness of estimated 3D edges. 
While the off-the-shelf 2D edge maps make for a relevant supervisory signal, we observe that they can introduce bias in the training or the evaluation, which calls for investigating better supervision in the future.

{\small
\PAR{Acknowledgements.}
This work was supported by 
the Czech Science Foundation (GACR) EXPRO (grant no. 23-07973X),
the Chalmers AI Research Center (CHAIR), WASP and SSF.
The compute and storage were partially supported by NAISS projects NAISS-2024/22-637 and NAISS-2024/22-237 respectively.
}

\clearpage
\appendix


\maketitlesupplementary

\cref{sec:add_imp} provides implementation details about the training of the edge-specialized Gaussian Splatting.
\cref{sec:qual} shows qualitative results over the scenes from the  Replica~\cite{replica19arxiv} dataset used by the authors of EMAP~\cite{Li2024CVPR} and three scenes from the Tanks and Temples dataset~\cite{Knapitsch2017}.
\cref{sec:limitations} discusses some limitations and failure cases of our method, pointing to relevant future work.

\section{Implementation details}\label{sec:add_imp}

\PAR{Initialization.}
\textit{Gaussian Position:}
For scenes from the DTU~\cite{jensen2014large}, Replica~\cite{replica19arxiv} and Tanks and Temples~\cite{Knapitsch2017} datasets, we use the SfM~\cite{schonberger2016structure} points as initialization.
Note that random point initialization also produces reasonable, but slightly worse results.
For \textit{ABC-NEF}~\cite{Ye_2023_CVPR}, we initialize our method with Gaussians centered at 10000 points randomly sampled in a unit cube.
This is because the dataset comprises texture-less objects for which \gls{sfm}~\cite{schonberger2016structure} generates extremely sparse or no point reconstruction at all.
\textit{Gaussian Scale}: We use a constant initial value of 0.004 %
for all datasets. However, a point-dependent value based on the complexity of the neighboring region may be more robust. 
\textit{Gaussian Opacity}: We use a constant initial value of 0.08 %
for all datasets.
\textit{Gaussian Orientation}: Random unit quaternions are used as initial values for all Gaussians.

\PAR{Training.}
We train the model for 500 epochs.
For the first 30 epochs, we only train the position parameters so that the scale and the orientation of the Gaussian do not compensate for its incorrect position during rendering.
Thus the training constrains the Gaussian's position, \ie, its mean, to lie on 3D edges.
We cull the Gaussians based on opacity and duplicate the ones with high positional gradients at regular intervals as in the original work~\cite{kerbl3Dgaussians}.
The learning rates of different parameters are as follows.
Position: starting with $1e^{-3}$, scaled with a factor of 0.75 every 10 epochs, 5 times.
Scale: $2e^{-4}$ constant.
Opacity: $3e^{-2}$ constant.
Orientation: $1e^{-3}$ constant.

We use $k=4$ as the number of nearest neighbors for computing $\mathcal{L}_{\text{orient}}$ defined in Eq.(4) of the main paper.
The weights of the loss function in Eq.(6) of the main paper are
$\lambda_{1} = 0.1$ and $\lambda_{2} = 0.1$ for object level scenes from 
\textit{ABC-NEF}~\cite{Ye_2023_CVPR} and DTU~\cite{jensen2014large}, while for larger scenes we use smaller values of $\lambda_{1} = 0.01$ and $\lambda_{2} = 0.01$.
The geometric regularization assumes that the Gaussians are already positioned close to the edges, therefore we start applying this regularization at epoch 300.
Note that the computation of nearest neighbors, required for the geometric regularization is computationally intensive and we observe that it is sufficient to only apply this regularization once in every 10 steps of the training process.

\PAR{Clustering}
During clustering, the alignment threshold is $\theta = 0.8$ on \textit{ABC-NEF}~\cite{Ye_2023_CVPR}, which have clean straight lines, and $\theta = 0.6$ on DTU~\cite{jensen2014large} to account for the higher curvature of the shapes.
During the parametric edge fitting, we fit a curve whenever the curve residual error is $\delta = 0.5$ lower than the line residual error. For objects from the DTU dataset~\cite{jensen2014large}, due to the prior knowledge that the objects have more curves than lines, a larger $\delta = 1$ is used.

For any further clarifications, please refer to the code released at \href{https://github.com/kunalchelani/EdgeGaussians.}{https://github.com/kunalchelani/EdgeGaussians.}

\section{Additional Qualitative Results}\label{sec:qual}
\begin{figure*}
    \centering
    \includegraphics[width=\linewidth]{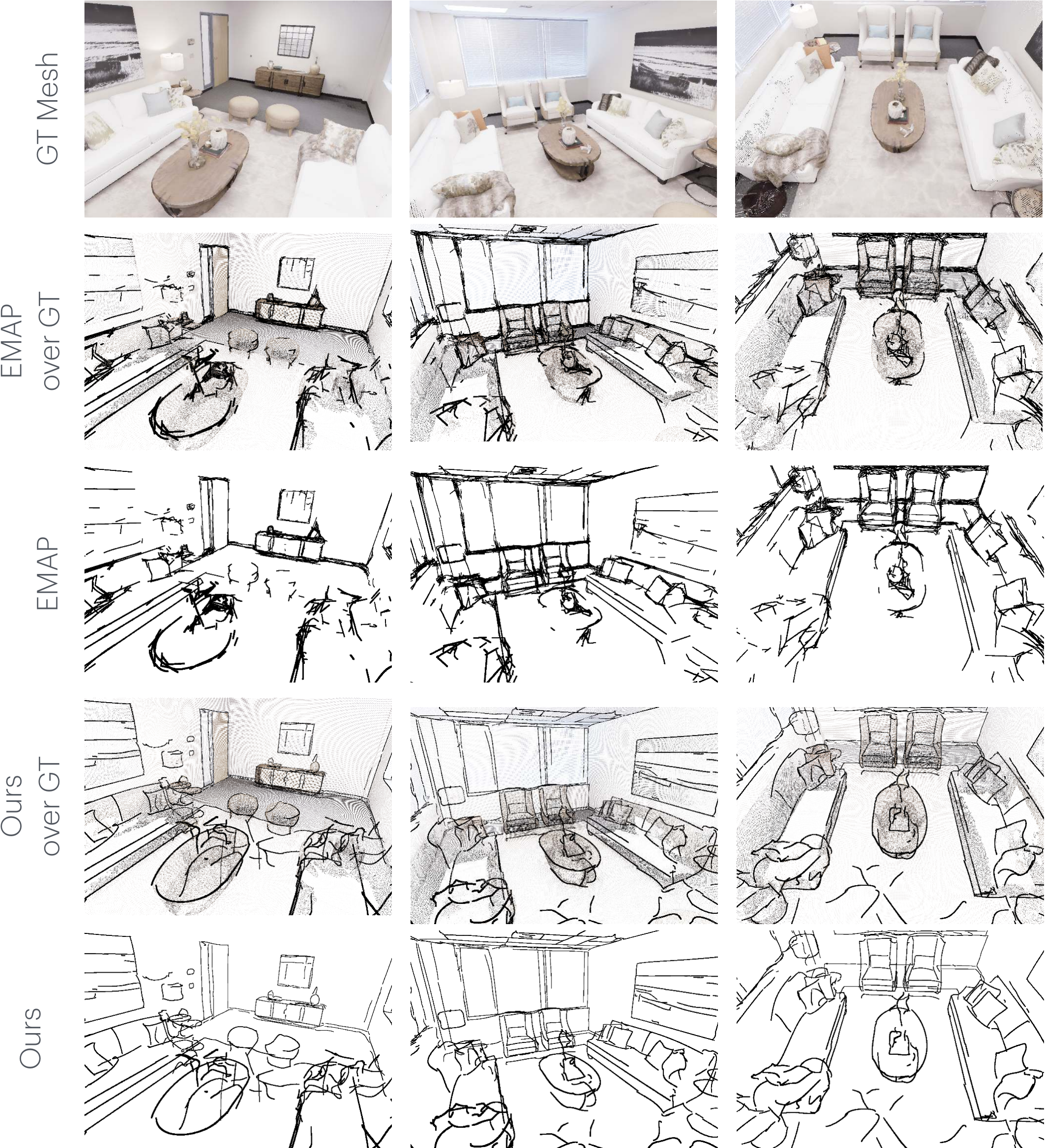}
    \caption{\textbf{Replica~\cite{replica19arxiv} \textit{room\_0}} : Qualitative result showing edges produced by our method and EMAP~\cite{Li2024CVPR}. In general it can be observed that EMAP~\cite{Li2024CVPR} has several duplicate / dense sets of edges close to ground-truth edges whereas our method produces clean single edges.}
    \label{fig:qual_replica0}
\end{figure*}

\begin{figure*}
    \centering
    \includegraphics[width=\linewidth]{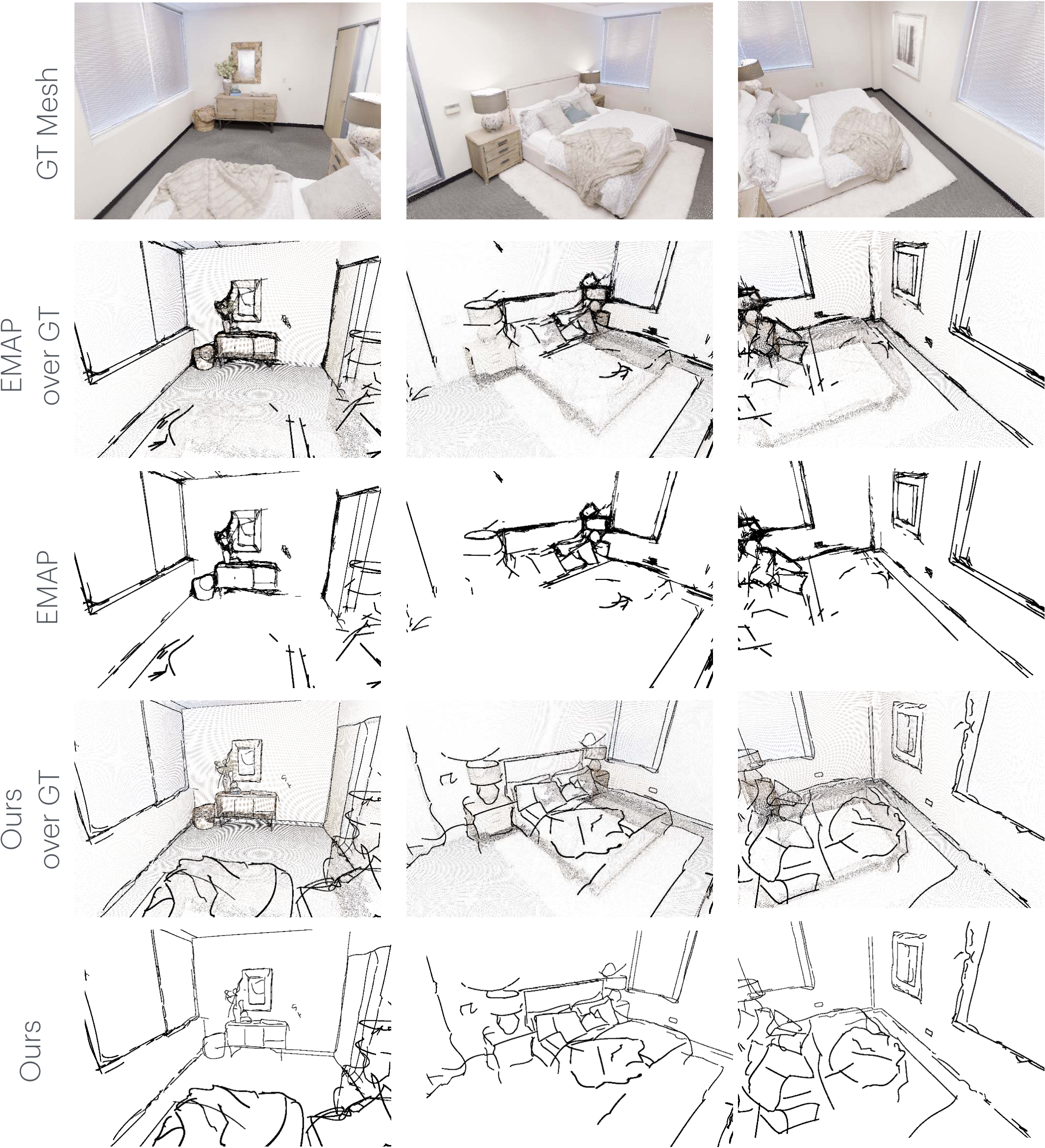}
    \caption{\textbf{Replica~\cite{replica19arxiv} \textit{room\_1}} : Qualitative result showing three different views of edges produced by our method and EMAP~\cite{Li2024CVPR}. In general it can be observed that EMAP~\cite{Li2024CVPR} has several duplicate / dense sets of edges close to ground-truth edges, while our method produces clean single edges.
    However, EMAP~\cite{Li2024CVPR} produces more accurate lines for some geometric edges, for example, on the window pane.}
    \label{fig:qual_replica1}
\end{figure*}

\begin{figure*}
    \centering
    \includegraphics[width=\linewidth]{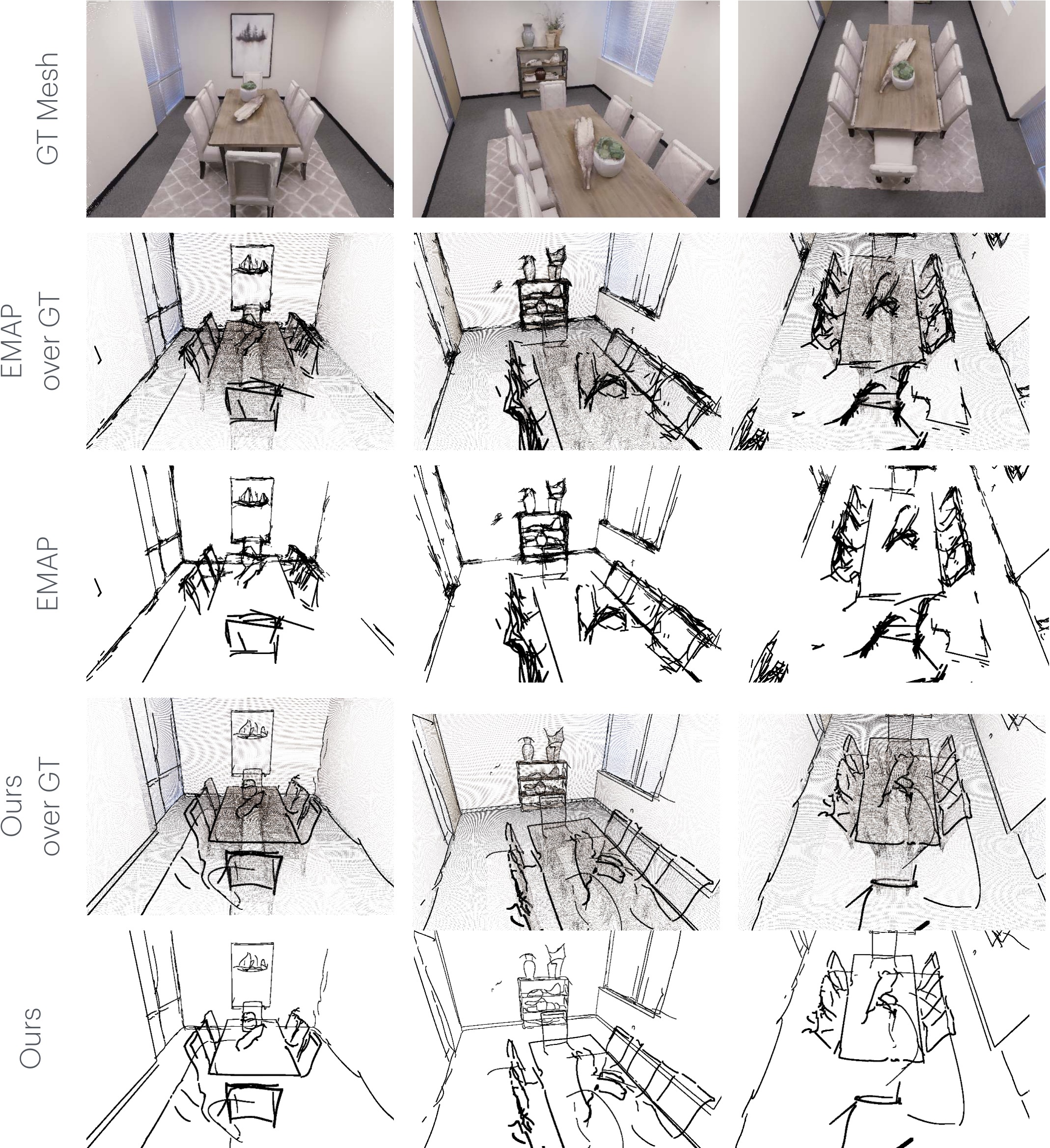}
    \caption{\textbf{Replica~\cite{replica19arxiv} \textit{room\_2}} : Qualitative result showing edges produced by our method and EMAP~\cite{Li2024CVPR}. In general it can be observed that EMAP~\cite{Li2024CVPR} has several duplicate / dense sets of edges close to ground-truth edges, while our method produces clean single edges.}
    \label{fig:qual_replica2}
\end{figure*}

\cref{fig:qual_replica0} to~\ref{fig:qual_replica2} show the results for the scenes \textit{room\_0}, \textit{room\_1}, \textit{room\_2} of the Replica~\cite{replica19arxiv} dataset. 
The results show that our method produces edges with a higher completeness than EMAP~\cite{Li2024CVPR}. 
Also, EMAP~\cite{Li2024CVPR} predicts clusters of duplicate edges close to the ground-truth edge, which is not a desirable result 
as it makes the reconstruction less sharp.
Overall, our method produces clean single edges that are more complete.
However, in some cases, EMAP~\cite{Li2024CVPR} produces geometrically accurate lines, which our method captures as incomplete curves.
\cref{fig:qual_replica1} shows one such example.
Although this can be partially addressed by adjusting the parameter $\delta$ involved in the model selection when fitting a line or a curve, this could be seen as a current limitation of our method.

\section{Limitations and Failure Cases}\label{sec:limitations}
\begin{figure*}
    \centering
    \includegraphics[width=0.8\linewidth]{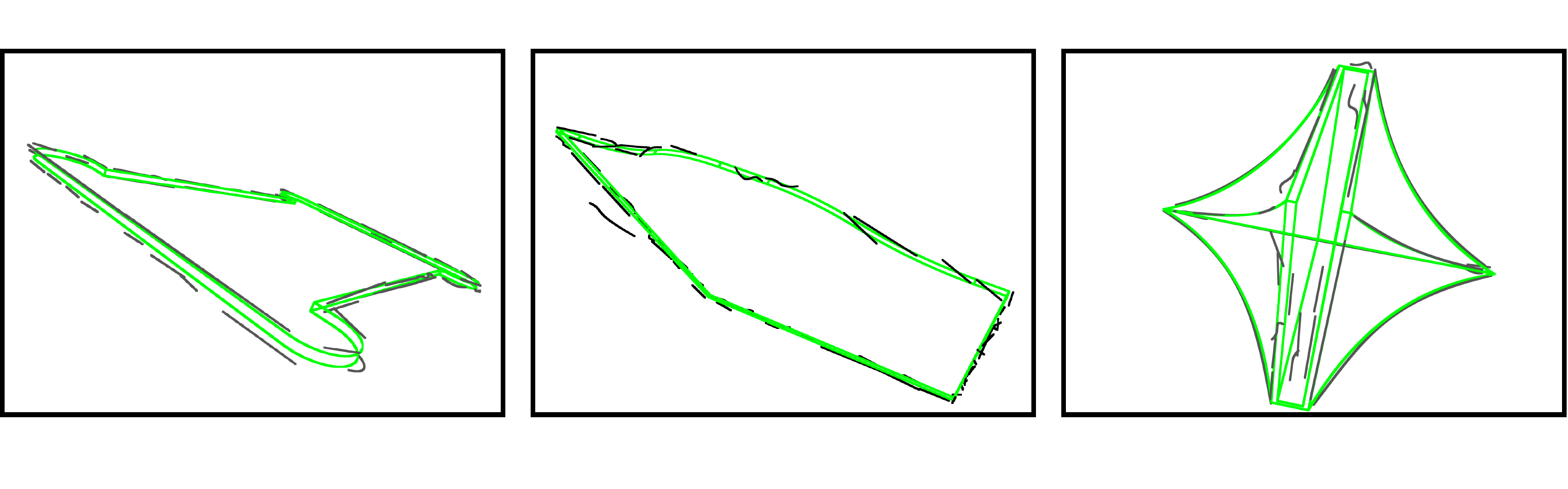}
    \caption{\textbf{Failure cases} : Scans \textit{00009685}, \textit{00002412} and \textit{00003884} (left to right) on the \textit{ABC-NEF}~\cite{Ye_2023_CVPR}. 
    The edges predicted by our method are shown in black and the ground-truth ones in green.
    These examples are challenging because they show extremely thin structures: the projection on two distinct parallel and close 3D edges can get projected into a single edge in several views of the supervisory 2D edge maps~\cite{su2021pixel,poma2020dense}.
    Another example where the proposed method is incomplete (right) is when the object has 3D edges inside the structure that are not detected by the 2D edge detectors.
    Then, there is no supervisory signal for those 3D edges.
    }
    \label{fig:failure}
\end{figure*}

\PAR{Fine structures and geometric regularization.}
As briefly described in the main paper, 
the method is limited by the noise in the supervisory signal of the 2D edge maps.
In many cases the fine structures in such edge maps~\cite{su2021pixel,poma2020dense} are not discernible, leading to incorrectly positioned edge points. 
Geometric regularization applied to noisy edge points can lead the Gaussians' 
to form short local curves to satisfy the alignment with their nearest neighbors. Examples of such cases can be seen in~\cref{fig:failure}. 

\begin{figure*}
    \centering
    \includegraphics[width=\linewidth]{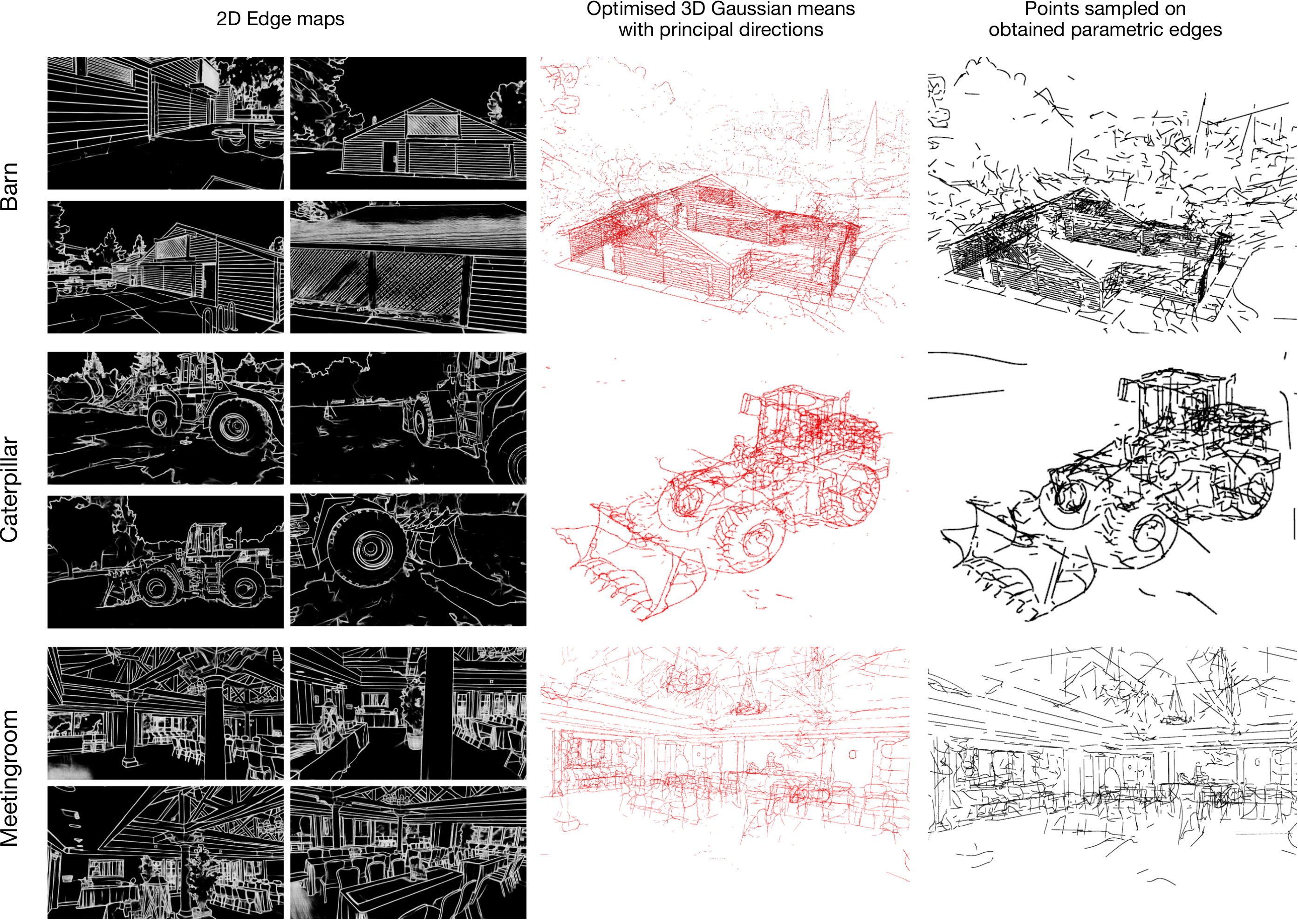}
    \caption{\textbf{Tanks and Temples~\cite{Knapitsch2017}}: Qualitative result showing edges produced by our method on three scenes from the tanks and temples dataset. Supervisory signal (Left), edge points represented as a small line segment centered at the mean of the optimized 3D Gaussians and oriented towards their principal directions (Middle) and the points sampled on the parametric 3D edges estimated (Right). Note that the estimated Gaussians faithfully represent the scene but the clustering and edge fitting process have room for improvement 
    as many correct edges are missed and spurious ones are created in this process.}
    \label{fig:qual_tnt}
\end{figure*}

\PAR{Clustering and edge fitting.}
Further, the clustering algorithm exhibits limitations when applied to larger scenes with complex structures. 
\cref{fig:qual_tnt} shows examples from the Tanks and Temples dataset~\cite{Knapitsch2017} where the oriented edge points (red), \ie, the 3D Gaussians, better cover the ground-truth 3D edges than the paramtric edges (black).
One explanation is that the graph traversal based clustering removes several correct edge components close to the true scene structure while including several incorrect edges. 
Instead of relying only on local geometric heuristics, defining a prior on which parts of the scene are more likely to hold 3D edge could improve the method's robustness.

\clearpage
\clearpage

{\small
\bibliographystyle{ieee_fullname}
\bibliography{egbib}
}

\end{document}

%% file: preamble.tex
\usepackage[dvipsnames]{xcolor}

\newcommand{\PAR}[1]{\noindent{\bf #1~}}

\usepackage{comment}
\usepackage[acronym]{glossaries}
\usepackage{multirow}

\makeglossaries{}
\newacronym{sota}{SotA}{State-of-the-Art}
\newacronym{nn}{NN}{Nearest-Neighbor}
\newacronym{sfm}{SfM}{Structure-from-Motion}
\newacronym{slam}{SLAM}{Simultaneous Localization And Mapping}
\newacronym{tab}{Tab.}{Table}
\newacronym{sdf}{SDF}{Signed-Distance-Function}
\newacronym{udf}{UDF}{Unsigned-Distance-Function}
\newacronym{gs}{GS}{Gaussian Splats}

\makeatletter
\DeclareRobustCommand\onedot{\futurelet\@let@token\@onedot}
\def\@onedot{\ifx\@let@token.\else.\null\fi\xspace}

\def\eg{\emph{e.g}.} 
\def\ie{\emph{i.e}.}

\makeatother